\pdfoutput=1
\documentclass[11pt]{article}

\usepackage{acl}

\usepackage{graphicx}
\graphicspath{ {./images/} }

\usepackage{times}
\usepackage{hyperref}
\usepackage{latexsym}
\usepackage{booktabs}
\usepackage{appendix}
\usepackage{titlesec}
\usepackage{float}
\usepackage{multirow}
\usepackage[T1]{fontenc}

\usepackage[utf8]{inputenc}

\usepackage{microtype}

\title{Perplexed by Quality: A Perplexity-based Method for Adult and Harmful Content Detection in Multilingual Heterogeneous Web Data}

\author{Tim Jansen, Yangling Tong, Victoria Zevallos, Pedro Ortiz Suarez \\
  Data and Web Science Group\\
  University of Mannheim \\
  \texttt{\{tijansen, ytong, lzevallo, portiz\}@mail.uni-mannheim.de}}

\begin{document}
\maketitle
\begin{abstract}
  As demand for large corpora increases with the size of current state-of-the-art language models, using web data as the main part of the pre-training corpus for these models has become a ubiquitous practice. This, in turn, has introduced an important challenge for NLP practitioners, as they are now confronted with the task of developing highly optimized models and pipelines for pre-processing large quantities of textual data, which implies, effectively classifying and filtering multilingual, heterogeneous and noisy data, at web scale. One of the main components of this pre-processing step for the pre-training corpora of large language models, is the removal of adult and harmful content. In this paper we explore different methods for detecting adult and harmful of content in multilingual heterogeneous web data. We first show how traditional methods in harmful content detection, that seemingly perform quite well in small and specialized datasets quickly break down when confronted with heterogeneous noisy web data. We then resort to using a perplexity based approach but with a twist: Instead of using a so-called \emph{``clean''} corpus to train a small language model and then use perplexity so select the documents with low perplexity, i.e., the documents that resemble this so-called \emph{``clean''} corpus the most. We train solely with adult and harmful textual data, and then select the documents having a perplexity value above a given threshold. This approach will virtually cluster our documents into two distinct groups, which will greatly facilitate the choice of the threshold for the perplexity and will also allow us to obtain higher precision than with the traditional classification methods for detecting adult and harmful content.
\end{abstract}

\section{Introduction}

Various interesting topics have emerged with the development of Natural Language Processing (NLP) in recent years. More and more popular are also topics of text classification in connection with a semantic analysis of the provided texts. Today's social developments, such as the recurring occurrence of hate speech, adult and harmful content in our current web structure, also influence the development of NLP.

For this reason, \citet{hateXplain} and many others deal with the automatic discovery and classification of this content. The extent to which state-of-the-art models, such as those presented by \citet{asogwa2022hate} and \citet{DeepLearningForHateSpeechDetection} in their publications, can be successfully applied to a large corpus crawled from the web in the multilingual domain will be explored and presented experimentally.

In particular, the OSCAR (Open Super-large Crawled Aggregated coRpus) corpus was presented in its first version by \citet{OrtizSuarezSagotRomary2019,ortiz-suarez-etal-2020-monolingual} and in its latest version, by \citet{OrtizTowards} contains filtered and language-sorted data from the Common Crawl and makes it available for application to NLP tasks. In order to provide a good data structure for possible NLP models pre-trained with OSCAR data, harmful content should be filtered out if necessary. Therefore, this publication aims to present feasible ways to annotate and filter this content from the OSCAR corpus. Besides the performance of the classification, the experiments also focus on its effectiveness and speed.

Our paper has five sections. In the first one, we presented an overview of related works in the research field of hate speech detection and classification of harmful content. In the second one, we described the structure of the datasets used to train the classifiers and language models. Then on the third one, we described in detail the implementation of three different approaches to identify harmful content in the OSCAR Corpus. For each approach, we describe the pre-processing steps, the model selection and configuration, and the final implementation to predict harmful content on a subset of OSCAR Corpus. Afterwards, we discuss and compare the results of the three approaches and identify the challenges and limitations encountered.

\section{Related Work}
Detecting hate speech, harmful and adult content has gained enormous importance in recent years. Due to the establishment of all social media platforms, the spread of this content has become much more prevalent. Comments with a negative and often offensive background are the order of the day and are beginning to have an extreme impact on today's society.

For this reason, there are already some promising models and experiments that aimed at detecting hate speech and harmful content. Models based on statistical machine learning, deep neural architecture, like CNN and LSTM, or the popular transformer-based methods, show promising results \citep{DeepLearningForHateSpeechDetection}.

The work of \citet{DetectingHarmfulOnlineConversationalContenttowardsLGBTQIA+Individuals} and \citet{fairlyaccurate} use machine learning methods such as logistic regression and support vector machines with harmful text. \citet{DetectingHarmfulOnlineConversationalContenttowardsLGBTQIA+Individuals} for example use a dataset based on Reddit conversations to detect LGBTQIA+ hostile content. How far their solution can be used to the differentiated data and its amount in the OSCAR Corpus will be empirically explored.

In contrast to the previously mentioned publications, \citet{ExploringHateSpeechDetectionWithHateXplainAndBert} use a transformer-based approach that uses the BERT model, well known in NLP. They use a dataset that contains a multiclass classification of hate and offensive language. The transformer-based BERT approach \citep{devlin-etal-2019-bert} will also be used in our implementation, and its feasibility will be tested.

Further groundwork is laid by \citet{DeepLearningForHateSpeechDetection}, who uses both methods mentioned above to detect hate speech. Besides the models' performance in classification, they explore their effectiveness and performance on three differentiated datasets. The effectiveness and performance of the methods used are also of great concern in the context of this project.

A method that is gaining traction to detect \emph{low-quality} documents, and in particular, documents containing hate speech, harmful and adult content in large crawled-corpora is the one introduced by \citet{wenzek2019monolingual}, where lightweight language models \citep{heafield2013kenlm} are trained with \emph{``clean and high quality''} corpora such as Wikipedia and then a perplexity-based filtration approach is used where on removes high-perplexity documents. This method was more recently used in the filtration pipeline of the BigScience ROOTS corpus \citep{laurencon-2022-roots} used to pre-train the multilingual BLOOM model \citep{bigscience-2022-bloom}.

\section{Data}
In total, we use four primary datasets, and they are used in the three different approaches. The datasets are the following:

\textbf{Hate Speech from Twitter}
A dataset for Twitter\footnote{Available on: \url{https://github.com/sharmaroshan/Twitter-Sentiment-Analysis/}} was used in the first approach for detecting hate speech in tweets. There are two attributes: tweet and label. The tweet contains the content of the text, and the label attribute is the annotation of the tweet. There are two classes - 1 and 0, where label '1' implies the tweet is hate speech, sexist or racist content, and label '0' represents the tweet is not hate speech. The distribution of both classes is not balanced. The data contains 31962 examples, which include 29720 not hate speech and 2242 hate speech, corresponding to an approximate ratio of 13:1.

\textbf{Mixed Data from Twitter}
In order to balance the first dataset, we combined it with another cyberbullying Twitter dataset\footnote{Available on: \url{https://www.kaggle.com/datasets/andrewmvd/cyberbullying-classification}}. The combined Twitter dataset was used in the second try of the first approach to detect harmful content. This cyberbullying dataset contains more than 47,000 tweets annotated according to 5 categories: Age, Ethnicity, Gender, Religion, Other types of cyberbullying, and Not cyberbullying. We converted all the cyberbullying labels to '1' and Not cyberbullying to '0.' Then we combined the first hate speech dataset with the cyberbullying dataset. The combined dataset contains 79654 examples in total, which includes 37665 not harmful content and 41989 harmful content, with an approximate ratio of 1: 1.11.

\textbf{Mixed Data from OSCAR \& The Pile}
For the second approach, we created a dataset from only adult content from OSCAR 22.01 \citep{OrtizTowards} combined with non-harmful data coming from the non-crawled part of The Pile corpus \citep{gao-etal-2020-pile}. The corpus is distributed in JSONLines format \citep{AbadjiOrtizSuarezRomaryetal2021}, meaning each line represents a single document encoded in a JSON object. We extracted 3292 rows from the English corpus containing the annotation of 'adult'. In order to generalize the dataset, we extracted 3420 rows of non-harmful English data from 20 sources coming from The Pile corpus, including Wikipedia, YouTube Subtitles, Enron Emails, etc. The distribution of harmful and non-harmful data is balanced.

\textbf{Adult Data from OSCAR}
For the third approach, we created two datasets. The first one is a small set for the proof of concept and the second one is a larger version of it. For the small harmful OSCAR dataset, the training set contains 2634 harmful content, and the validation set and test set both contain 829 examples of harmful content from OSCAR and non-harmful content from the same resources as the previous dataset. After establishing the utility of the data for the third approach, a larger harmful OSCAR dataset was generated. For its creation, we extracted 23702 harmful content from the first 900 English OSCAR files as the training set, and the validation and test set both contain 4673 harmful and non-harmful examples. The ratio of the non-harmful and harmful content on these sets is 63:37. The training dataset is 136 MB, the validation set is 274 MB, and the test dataset is 143 MB. The reason that the size of the training data is not the biggest is that the testing and validation set contains non-harmful content from The Pile.

\section{Methodology}
As previously mentioned, our main objective is to propose an efficient methodology for labeling harmful content from large volumes of multilingual data extracted from the Internet. To accomplish this challenge, we experimented with three approaches to find the most suitable solution for our task.

In the first one, the problem is posed as a classification task. The idea is to implement machine learning and deep learning models trained on annotated twitter datasets to predict harmful content in the OSCAR dataset. In this context, Twitter's Hate Speech and Twitter's Mixed Data datasets were used as human experts annotated the data. Nevertheless, since the structure of such tweet-based datasets differs from text extracted from web pages, another approach was proposed to overcome this limitation.

The second approach emerged as an alternative to the first one to train the more classifiers from the previous approach with the Mixed Data from OSCAR \& The Pile, whose structure is similar to OSCAR's dataset and, therefore, a setting closer to real-world web data is expected.

Finally, our third approach tackled the problem from another angle, based on language models and perplexity. For this purpose, we estimated unpruned language models of the adult data extracted from the OSCAR corpus using modified Kneser-Ney smoothing \cite{heafield2013kenlm}. Afterwards, we calculated the perplexity of the model on harmful and non-harmful content, compared these results, determined a threshold and used it to determine at which perplexity value a text will be considered harmful. Finally, this procedure was applied to the OSCAR corpus to label new harmful content.

\subsection{First approach}
The first approach uses the text classification technique and tries to recognize the semantics of the texts. Already mentioned basics in the detection of hate speech or harmful content of different backgrounds are also based on this attempt, which is why the hypothesis exists that this approach can be successful. Using Twitter data containing annotated hate speech, classical machine learning models, the FastText classifier and transformer models will be trained and tested for their performance and usability. After training the classification models, the first approach will also include an attempt to classify OSCAR data that is predicted using the created models.

\begin{table*}[!ht]
  \centering\small
  \begin{tabular}{@{}lccc@{}}
    \toprule
    Model                          & Macro F1 & Training Speed (s) & Testing Speed (s) \\ \midrule
    Naïve Bayes                    & 85\%     & 3.33               & 0.103             \\
    Random Forest                  & 88\%     & 11054.18           & 5.807             \\
    Logistic Regression            & 89\%     & 27.54              & 0.094             \\
    SVM                            & 89\%     & 11.94              & 0.111             \\
    SGD                            & 90\%     & 8.13               & 0.094             \\
    FastText                       & 89\%     & 3.41               & 0.764             \\
    DistilBERT (trained on GPU)    & 91\%     & 2783               & 21.21             \\
    DistilRoBERTa (trained on GPU) & 92\%     & 3127               & 23.51             \\ \bottomrule
  \end{tabular}
  \caption{Models performance on the Mixed Twitter data with the best preprocessing steps
  }
  \label{preprocess}
\end{table*}

\subsubsection{Classical models}
Eight different classical machine learning supervised models were selected in terms of performance and variety. We trained all models on the Hate Speech Twitter dataset in the first run with the initial preprocessing steps. The models were trained on the more extensive combined Twitter dataset with the best preprocessing steps to optimize the performance.
\paragraph{Data Preprocessing} \hfill\break
In order to obtain high-performance classical models, the input data must be optimally preprocessed. To find the  best mixture of preprocessing steps the Mixed Data from Twitter was used. Different combinations were tested with the help of the NLTK and the spaCy frameworks. Lowercasing the text data and removing special characters and URLs did not yield any measurable success in terms of model performance. An initial improvement in performance was achieved by tokenizing the text and removing the existing stopwords. The two frameworks tested the combination of several tokenizers and stopword lists. The combination of the spaCy framework's stopword list and the word tokenizer's use from NLTK turned out to be the most successful. For further improvement, mechanisms of stemming and lemmatization were applied. The biggest enhancement was achieved by adding the Word Lemmatizer from NLTK. As the last step, the replacement of emojis was added to the preprocessing, which increased the classification performance as measured by the F1 macro score. Thus, the best preprocessing pipeline comprises tokenization, stopword removal, lemmatization and the replacement of the emojis. In the following experiments, as the results of Table \ref{preprocess} show, this preprocessing pipeline was applied.

\paragraph{Model Selection}\hfill \break
On the first try, we trained eight classical models to get a baseline and compare the performance of different models. In the second try of the classical models, we selected the best five classical models in terms of the variety, performance, and speed to be trained on the Mixed Twitter data. The models are Naïve Bayes, KNN, Decision trees, Extra Trees, Random Forest, Logistic regression, SGD and SVM.

\paragraph{Training}\hfill \break
For classical models, we needed to train them from scratch. We vectorized the data using scikit-learn TF-IDF, and the parameters included max\_df, smooth\_idf, and norm method. We trained the models with pipeline and GridSearchCV functions according to different classifiers. Finally, the best parameters were used on the predicting dataset.

\subsubsection{Transformer models}
One of the main disadvantages of classical models is that they highly rely on pre-processing steps, as we demonstrated. Then a lot of computational power is required to find the right steps for the task and the target language. Despite this, there is no guarantee of finding sensible features for the task. Furthermore, a text structure is complex and has semantic meaning; therefore, to classify harmful content, we need a model that can better model the use of language in a highly heterogeneous context. For this we use transformer-based architectures \citep{Vaswani2017} that have proved highly effective in a wide range of tasks in recent years, even in environments with little to no training data \citep{NEURIPS2020_1457c0d6}.

\paragraph{Model Selection}\hfill \break
To overcome some of the disadvantages of classical machine learning models, we decided to fine-tune two of the most popular transformer architectures, BERT \citep{devlin-etal-2019-bert} and RoBERTa \citep{liu2019roberta}. As the computational cost and hardware requirements to fine-tune and run these models over vasts amounts of data are quite high, and since the infrastructure that we have access to is quite limited, we use their distilled variants distilBERT and distilRoBERTa \citep{sahn2019distilbert} instead of using the original BERT and RoBERTa models.

\paragraph{Training}\hfill \break
The distilBERT and distilRoBERTa pre-trained architectures are fine-tuned using the python Flair library. This is a simple, state-of-the-art NLP framework built directly on top of PyTorch \cite{flair}. No data pre-processing was necessary, as both models generate embeddings from the raw data. We fine-tune pre-trained base versions, on the one hand, with Twitter's Hate Speech dataset and, on the other hand, with Twitter's Mixed Data dataset. The models were trained for seven epochs, but the model that best generalizes unseen data, with no overfitting, was found in the second epoch in both cases. To find the best values for the hyperparameter's learning rate and minimum batch size, we used the random search method and found that, for both models, a learning rate of 5.0e-5 and a mini-batch of 4 give the best performance. The results of the models are shown in Table \ref{preprocess}.

\subsubsection{FastText Model}
Both the classical models, which require enormous computation to perform the associated preprocessing, and the presented deep learning models, which usually offer outstanding performance but require a lot of computation and time, have disadvantages. The above arguments must be harmonious to represent a reasonable solution for classifying a large text corpus. This is the motivation for applying another classifier, which combines the mentioned points and is presented in the following.

The FastText classifier is a method presented by \citet{BagOfTricksforEfficientTextClassification} in their publication for fast and simple classification of large amounts of text data. During the development of the classifier, the focus was on the fast applicability to large text datasets, which should be classifiable within a few minutes \citep{BagOfTricksforEfficientTextClassification}.

\begin{table*}[!ht]
  \centering\small
  \begin{tabular}{@{}lllllll@{}}
    \toprule
    \multirow{5}{*}{Model} & \multicolumn{6}{c}{Macro F1}                                                                                                                                                                                                                                                                                                         \\ \cmidrule(l){2-7}
                           & KNN                          & \begin{tabular}[c]{@{}l@{}}Naïve \\ Bayes\end{tabular} & \begin{tabular}[c]{@{}l@{}}Decision \\ Tree\end{tabular} & \begin{tabular}[c]{@{}l@{}}Random \\ Forest\end{tabular} & \begin{tabular}[c]{@{}l@{}}Logistic \\ Regression\end{tabular} & \begin{tabular}[c]{@{}l@{}}Extra \\ Tree\end{tabular} \\
                           & 64\%                         & 68\%                                                   & 70\%                                                     & 73\%                                                     & 75\%                                                           & 77\%                                                  \\ \cmidrule(l){2-7}
                           & SGD                          & SVM                                                    & FastText                                                 & DistilBert                                               & DistilRoberta                                                  &                                                       \\
                           & 77\%                         & 78\%                                                   & 82\%                                                     & 86\%                                                     & 86\%                                                           &                                                       \\ \bottomrule
  \end{tabular}
  \caption{Models performance on the Hate Speech dataset with initial preprocessing steps}
  \label{initail}
\end{table*}

\paragraph{Data Preprocessing}\hfill \break
The FastText classifier was initially used without preprocessing. As seen in table \ref{initail}, very good results were already achieved with the associated data set. Different preprocessing steps were experimented with to improve the performance concerning the creation of the word representations. The same steps and tries in comparison to the Classical Models were executed. The fewest steps could improve the F1 Macro Score using the Mixed Data from Twitter dataset. The best preprocessing was found only using the step of Lemmatization using the word tokenizer out of the NLTK python framework and the WordNetLemmatizer also coming from the NLTK framework. Table \ref{preprocess} shows the results with applied preprocessing and the improved dataset of the first approach. The F1 macro score improved compared to the trial without preprocessing the initial dataset from 82\% to 89\%.

\paragraph{Training}\hfill \break
The training of the FastText model can be done quickly due to the architecture of the classifier. The training was performed with 10 epochs and 4 threads without hyperparameter tuning. In this way, good performance was already achieved in a very good training time of 3.41 seconds using the Mixed Data from Twitter, as seen in Table \ref{preprocess}. The framework provides an autotune option to perform automatic hyperparameter tuning. Performing auto-tuning significantly increases training time without significantly improving the model performance. For this reason, the initial configuration was preferred.

\begin{table*}[!ht]
  \centering\small
  \begin{tabular}{lrrrr}
    \toprule
    \multicolumn{1}{c}{\multirow{2}{*}{Approach}}     & \multicolumn{1}{c}{\multirow{2}{*}{Model}} & \multicolumn{1}{c}{\multirow{2}{*}{\begin{tabular}[c]{@{}c@{}}Macro F1 on\\  Test\end{tabular}}} & \multicolumn{2}{c}{Model Prediction on 1GB OSCAR}                                                                                             \\ \cline{4-5}
    \multicolumn{1}{c}{}                              & \multicolumn{1}{c}{}                       & \multicolumn{1}{c}{}                                                                             & \begin{tabular}[c]{@{}c@{}}Prediction \\ Speed (s)\end{tabular} & \begin{tabular}[c]{@{}c@{}}\% of  Harmful \\ Content Predicted\end{tabular} \\ \hline
    \multicolumn{1}{c}{\multirow{2}{*}{1st approach}} & \multicolumn{1}{c}{distilBERT}             & \multicolumn{1}{c}{91\%}                                                                         & 23529.01                                                        & 78.7\%                                                                      \\
    \multicolumn{1}{c}{}                              & \multicolumn{1}{c}{FastText}               & \multicolumn{1}{c}{89\%}                                                                         & 41.66                                                           & 74.8\%                                                                      \\ \hline
    \multicolumn{1}{c}{2nd approach}                  & \multicolumn{1}{c}{FastText}               & \multicolumn{1}{c}{91\%}                                                                         & 44.26                                                           & 65.4\%                                                                      \\ \hline
    \multicolumn{1}{c}{3rd approach}                  & \multicolumn{1}{c}{Perplexity\_4.22}       & \multicolumn{1}{c}{94\%}                                                                         & $\sim$50                                                        & 0.49\%                                                                      \\
    \multicolumn{1}{c}{3rd approach}                  & \multicolumn{1}{c}{Perplexity\_5.31}       & \multicolumn{1}{c}{98\%}                                                                         & $\sim$50                                                        & 0.79\%                                                                      \\
    \multicolumn{1}{c}{3rd approach}                  & \multicolumn{1}{c}{Perplexity\_13.51}      & \multicolumn{1}{c}{99\%}                                                                         & $\sim$50                                                        & 1.01\%                                                                      \\ \bottomrule
                                                      &                                            &                                                                                                  &                                                                 &
  \end{tabular}
  \caption{Models performance (1st approach: 1 GB OSCAR Data; 2nd \& 3rd approach: Mixed Data from OSCAR \& The Pile) }
  \label{OSCAR_prediction}
\end{table*}

\subsubsection{Evaluation}
We trained the models first with the Hate Speech Twitter data. Macro F1 score was chosen as the performance metric because this method treats all classes equally regardless of their support values. Table 1 shows that among the Classical models, KNN had the worst performance, obtaining a 64\% Macro F1 score; in contrast, SVM had the highest Macro F1 score, 78\%, FastText achieved a F1 score of 82\%, DistilRoBerta had the second highest F1 score, 86.03\%, and DistilBert had the best performance with a F1 score of 86.03\%.

To improve the performance of models, we combined Hate Speech Twitter data with Cyberbully Twitter data to get a larger dataset. After optimizing the preprocessing steps, we selected five classical models with the best combination of speed and performance. Table \ref{preprocess} shows that all the models obtained better results with the larger dataset and better-preprocessing steps. For F1 score, SGD got the best result among classical machine learning models, with a score of 90\%, FastText got a result of 89\%, DistilBert and DistilRoberta obtained 91.27\% and 91.57\%, respectively. In terms of training speed, Naive Bayes and FastText were the fastest, taking only 3.33 and 3.41 seconds to train on the mixed Twitter dataset, respectively. Random Forest was the slowest; it took 11054.18 seconds to train the model. For testing speed, SGD and Logistic Regression were the quickest. The transformer architectures DistilBERT and DistilRoBERTa were the slowest.

At the end of the first approach, we tried to predict 1 GB of OSCAR data. Classical models took too long to be preprocessed and vectorized, and we ended the pre-processing step after the fourth day. FastText took 41.66 seconds, and DistilBert spent 23529.01 seconds. Table \ref{OSCAR_prediction} shows the distribution of predicting labels from DistilBERT. DistilBERT predicted 78.7\% of the samples to be harmful, while FastText predicted 74.8\%. Both models predicted a considerable amount of instances as harmful. These results were surprising and unrealistic. Even though the main idea of our project is to find a model that can label unseen instances from OSCAR, it is highly unlikely that most of them are harmful. In fact, in 1 GB of OSCAR data, only 23 instances are labelled as harmful; such a big difference is not a realistic result. In consequence, we can conclude that our models performed very poorly on the OSCAR corpus.

\subsection{Second approach}
The first approach was performed based on two Twitter datasets. After successfully training the models, the classification of the OSCAR data, as described before, was unsatisfactory. The distribution of classifications created from the models from the first approach does not represent the true distribution of OSCAR data.

In order to create a better-generalized model, the second approach should use a differentiated data source. The hypothesis is that the datasets used in the first approach to train the models were too different from the text in the OSCAR corpus. To change the data basis and find a structure similar to the OSCAR data, the newly constructed Mixed Data from OSCAR \& The Pile was used in this approach.

We considered that with a usable data source, a classification model could achieve promising results. Since the FastText model from the first approach achieves the best combination between performance and classification speed, it will be used in the second approach. The goal is to construct a target-oriented model for classifying OSCAR data by combining the model and the new dataset. The same preprocessing is used for the FastText model as in approach one.

\subsubsection{Evaluation}
The FastText model obtained a good classification performance again, with an F1 macro score of 91\%. Also, a good result has been achieved in terms of speed when classifying an unknown data set, as before with the first approach. One gigabyte of unseen OSCAR data is classified in approximately 44 seconds as Table \ref{OSCAR_prediction} shows. Unfortunately, the prediction distribution of the OSCAR data still is not satisfying. However, the FasText model trained on Mixed Data from OSCAR \& The Pile predicted around 10\% less content as harmful compared with the first approach.
Nevertheless, it still predicted a huge percentage of the OSCAR corpus as harmful content. Table \ref{OSCAR_prediction} shows that nearly 65\% of the data records were classified as harmful, and only 35\% of the data records represented the class of non-harmful data. Therefore, even when the second approach showed improvement still did not produce realistic results. To overcome this obstacle, our third approach would face the problem from a different perspective, using language models and perplexity as presented in the following section.

\subsection{Third approach}
So far, we have trained different classifiers on different datasets to be able to detect harmful content on the OSCAR corpus. However, even when the machine and deep learning models achieved a good performance on the test sets, they could not be generalized to the OSCAR corpus, as we saw in Table \ref{OSCAR_prediction}. Consequently, our last approach tackles the problem differently. It implements a solution based on language models and perplexity inspired by the work of \cite{wenzek2019monolingual}, where they used a similar combination to extract high-quality monolingual datasets from web crawl data.

\begin{figure*}[!ht]
  \centering
  \includegraphics[scale=0.56]{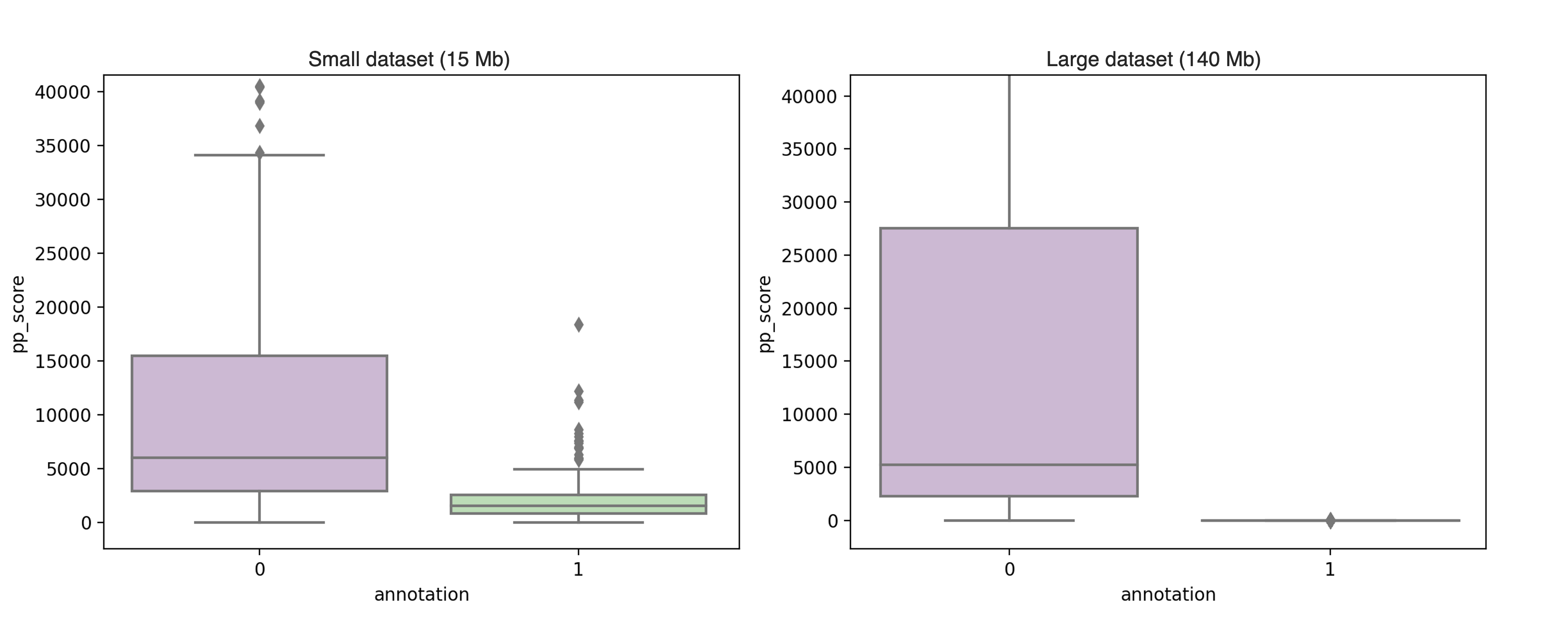}
  \caption{Perplexity Score Distribution}
  \label{3_perplexity_small_large}
\end{figure*}

\subsubsection{Model Selection}
As mentioned, the inspiration for this approach is the paper proposed by \citet{wenzek2019monolingual}. Here, the authors proposed a pipeline to extract monolingual datasets from web crawl data; their pipeline had the data processing steps introduced in FastText \cite{mikolov2017facebook} as a base, which then was augmented with a filtering step to select documents close to \emph{high-quality corpora} such as  Wikipedia \cite{wenzek2019monolingual}.  The filtering mechanism consists of training a language model on the targeted sources to use the perplexity as a quality scoring function for documents. The idea behind this filtering mechanism is what we used as the base for this approach.

We trained language models as in the \citet{wenzek2019monolingual} work. However, instead of using Wikipedia data for training the models and filtering the documents with high perplexity, we take the exact opposite approach: We trained an unpruned language model on the Harmful Data from OSCAR using modified Kneser-Ney smoothing. After training, this language model should have a high probability for the most frequent n-grams found in harmful content and a low probability for unseen sequences. Therefore, the model is expected to assign a low probability to a non-harmful text, translating into a high perplexity score. Under this assumption, we used the perplexity score of each document in the validation dataset composed of harmful and non-harmful content to determine the perplexity threshold to separate harmful and non-harmful content. We then validated the selected threshold on the test data. Finally, we applied this procedure to the documents in OSCAR's corpus to label them as harmful content if their perplexity score is below the chosen threshold.

\begin{figure*}[!ht]
  \centering
  \includegraphics[scale=0.56]{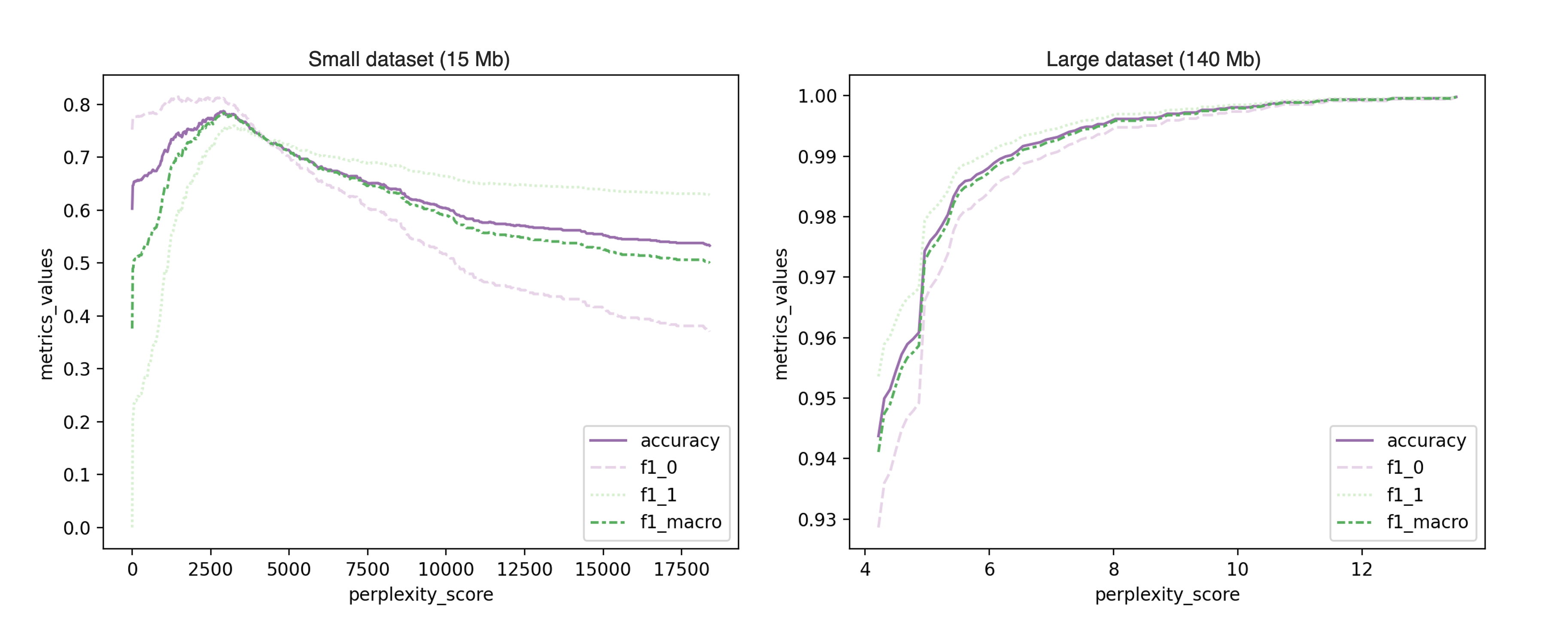}
  \caption{Classification Performance by Threshold on the Validation Set }
  \label{3_threshold_small_large}
\end{figure*}

\subsubsection{Training}

We used the modified Kneser-Ney smoothing implementation from the library KenLM \cite{heafield2013kenlm} to train a language model on OSCAR's Harmful Data. This implementation allowed efficient language model queries by reducing memory and time costs \cite{heafield2013kenlm}.

First, to test the concept, we used a small sample of harmful data to train the language model of about 14Mb, composed of 2634 documents. Once the model was trained, we applied it to the documents on the validation set, composed of harmful and non-harmful content, and calculated the perplexity score of the model for each document. A comparison of the resulting distributions of the perplexity score between harmful and non-harmful data is presented on the left-hand side of the figure \ref{3_perplexity_small_large}. Although the distributions overlapped slightly, it is clear that the first quartile (1Q) of the non-harmful content was higher than the third quartile (3Q) of the harmful content. This indicates that a good threshold should be between these two values. We calculated the model's performance with 100 different thresholds to select the best one. As the validation data was unbalanced, the classifier's performance was measured using the F1 macro score. On the left side of Figure \ref{3_threshold_small_large}, the model's performance according to the threshold employed was presented in terms of the metrics: F1-macro, F1-harmful, F1-non-harmful, and accuracy. The highest F1-macro value, 78.40\%, was reached when the perplexity threshold was 2906; after this point, the F1-macro score decreased, as well as the other metrics being F1-non-harmful, the most affected one.

Although we used a small dataset, this approach showed encouraging results. Therefore, to improve this method's performance and take advantage of the efficient KenLM implementation, we increased the size of the training dataset to 140Mb with 23702 instances. Then we trained another language model using this data and repeated the procedure explained in the previous paragraph. The comparison between the distributions of the perplexity score for non-harmful and harmful content is presented on the right side of Figure \ref{3_perplexity_small_large}. It is noticeable that there was almost no overlapping between both distributions. To find the appropriate threshold according to the new validation data, we plotted the same four metrics used previously for the possible thresholds on the right side of \ref{3_threshold_small_large}. This time the highest F1-macro value, 99.97\%, was reached when the perplexity threshold was 13.51, much lower than the one we had with the smaller dataset. This result shows that with more data, the model can better recognize harmful content reflected in a low perplexity value, while when faced with no-harmful content, it gets perplexed. Therefore, the perplexity score of the non-harmful text has a high value.

Even though it is clear that 13.51 was the best threshold value we could find based on the validation data, we had to make sure that the model could generalize well to the unseen data. Therefore, we will select two additional thresholds to compare the performance achieved by these three models on the test data. These additional thresholds were 4.22 and 5.31. The former is the maximum value of the harmful content, and the latter is the one that leads to the steepest step in the four metric curves.

\subsubsection{Evaluation}
We used the test data set to evaluate the performance of the three models (threshold = 4.22, 5.31 and 13.51) on unseen data. Similar to the estimation for validation, we applied the trained linguistic model to the documents in the test set and calculated the model perplexity for each document. We then used the three thresholds to classify a text as harmful if its perplexity value was below the threshold and otherwise non-harmful. We then compared these estimations with the actual labels and computed the classification metrics of the models. Since the test set was also unbalanced, we used the F1 macro score to compare the performance. The results are presented in Table \ref{perplexitythresholdsF1}. Pleasingly, the performance of the three models on previously unseen data was similar to their performance on the validation data. The model with a threshold = 13.51 maintained the best performance, with an F1 macro equal to 99\%. The other two models also obtained good F1 macro scores. This result indicates that the threshold selection did not overfit the validation data and could generalize to unseen data.

\begin{table*}[!ht]
  \centering\small
  \begin{tabular}{lrrrr}
    \toprule
    \multirow{2}{*}{\begin{tabular}[c]{@{}c@{}}Perplexity\\ Threshold\end{tabular}} & Validation & \multicolumn{3}{c}{Test}                               \\ \cmidrule(lr){2-2}  \cmidrule(lr){3-5}
                                                                                    & F1\_macro  & F1\_harmful              & F1\_no\_harmful & F1\_macro \\ \midrule
    4.22                                                                            & 94\%       & 95\%                     & 93\%            & 94\%      \\
    5.31                                                                            & 98\%       & 98\%                     & 97\%            & 98\%      \\
    13.51                                                                           & 99\%       & 100\%                    & 99\%            & 99\%      \\ \bottomrule
  \end{tabular}
  \caption{Comparison between the performance of the thresholds in the validation and test set}
  \label{perplexitythresholdsF1}
\end{table*}

We used our three promising models to predict the label of 1 GB of OSCAR corpus, and the prediction took around 50 seconds. The distribution of the predicted labels of the three models and prediction time is presented in Table \ref{OSCAR_prediction}. In contrast to the results of the first and second approaches, these three models predicted that most data was not harmful and only a tiny percentage was harmful. Specifically, our best model, threshold 13.51, predicted only 1.01\% of the data as harmful content, while thresholds 5.31 and 4.22 predicted 0.79\% and 0.49\%, respectively. This result is more realistic and could imply that the third approach is the most suitable solution for our task.

Finally, we look at the predictions made by the best model, with a perplexity threshold of 13.52, on two subsets of 1 GB each extracted from the English OSCAR's corpus. The first one was part\_1; this was one of the files from which we extracted harmful content to train the language model. In contrast, the second one, part\_1500, was neither part of the training, validation, nor testing. In Table \ref{cm_part1_part1500}, we presented the performance metrics true positives (TP), true negatives (TN), false positives (FP) and false negatives (FN) resulting from the comparison of the existing harmful label in the OSCAR corpus with the prediction of our model. For both data subsets, we see that there are instances belonging to the FN, which shows that the model could identify a reasonable amount of new harmful text that previously was not labelled as harmful. TP results implied that the model could label actual harmful content as harmful for the subset part\_1 but almost none for the subset part\_1500.

\begin{table}[!ht]
  \centering\small
  \begin{tabular}{lrrrrr}
    \toprule
    Part       & Size   & TP & TN     & FP & FN            \\ \midrule
    part\_1    & 119325 & 28 & 118110 & 0  & \textbf{1187} \\
    part\_1500 & 119280 & 4  & 118024 & 27 & \textbf{1225} \\
    \bottomrule
  \end{tabular}
  \caption{Comparison of the metrics TP, TN, FP, and FN between part\_1 and part\_1500 of OSCAR corpus }
  \label{cm_part1_part1500}
\end{table}

These findings are further discussed in the following section, but generally, the results of the third approach are the most satisfactory. The model has a high performance not only in the validation but also in the test data. Additionally, prediction on the OSCAR corpus produced realistic results, unlike the two previous approaches. Moreover, due to the efficient implementation of KenLM, this technique can be used to predict the labels for the entire OSCAR corpus.

\section{Discussion}
The first approach trained several classical classifiers, transformer architectures, and FastText classifiers on Mixed Data from Twitter to detect harmful content on the OSCAR corpus. The performance of the models on the Twitter test set was satisfactory by enlarging the size of the dataset and optimizing the pre-processing steps. However, the distribution of predicting 1 GB OSCAR data was unrealistic, which predicted the majority of content to be harmful. The result could be explained by the fact that Twitter's data structure differs from the OSCAR corpus, and we used the Twitter data in the first place because it was hard to find harmful labeled data from other sources.

The limitations and results of the first approach resulted in a change in our course of action. As mentioned before, we assumed that the different data structure in the first approach was responsible for the unsatisfactory results of the predictions. To adapt the data structure of the training data of the FastText model to the data of the OSCAR corpus, we decided to create and use the Mixed Data from OSCAR \& The Pile dataset in the second approach. The hope was to obtain a model that better generalizes the data. The performance of the FastText model, in terms of the test set of the data we used, was very good, so we expected a better outcome. After we got the results of the prediction of the OSCAR data, we could see an improvement in the distribution. The reason for this seems to be the more similar data structure of the training data in comparison to the OSCAR data. Despite an improved result, we assume that the obtained distribution of harmful and non-harmful data still looked unrealistic since more than half of the instances were classified as harmful. All these results were obtained by using the FastText classifier alone. This model was chosen because of its performance and speed. Our computational resources and experience from the first approach made this selection reasonable. For further verification of the observed results, additional classification algorithms could be run with the necessary computational resources. However, it can be assumed that no evidence suggests that other models could result in a more realistic solution.

Our third approach did not tackle the problem by training a classifier but a language model from Harmful Data from OSCAR. We then used the perplexity score to determine whether a text was harmful or not. We selected this methodology as it was effectively used in other works on large datasets crawled from the internet. In the previous section, we also showed that increasing the training data was beneficial as it allowed the model to clearly distinguish between harmful and non-harmful text, increasing the model performance and even virtually clustering documents into harmful and not-harmful as we can see in Figure \ref{3_perplexity_small_large}. Unfortunately, creating a more extensive training dataset was a time-consuming process limited by the hardware configuration we used in our project. Nevertheless, like the two previous approaches, this one also performed well on the test data. However, unlike the others, it could identify a realistic amount of new harmful texts in the OSCAR corpus that better reflect the result of previous studies \citep{kreutzer-etal-2022-quality}. One pitfall of the model was that it did not assign the harmful label to a large percentage of entries pre-labelled as harmful in the files that were not used to create the Harmful Data from OSCAR dataset. This issue could occur because the Harmful Data from OSCAR dataset was constructed sequentially. Then, the model learned to identify the type of harmful content present in the first 900 files from the English section of the OSCAR corpus. A similar explanation can be that the content of the OSCAR data pre-labelled as harmful differs tremendously among them. Hence, as the model was trained only with a part of the files, it could not identify the type of harmful content presented in the other part. If this is the case, a solution would be to train the KenLM models with the entire adult-annotated data for each language. Table \ref{size_adult} in the Appendix, shows the sizes of adult-tagged data in OSCAR 22.01 by language, which from our results, also suggests that we could potentially use this method in 35 OSCAR sub-corpora.

A meaningful advantage of this approach was its fast prediction speed, making it feasible to apply it to large volumes of data and obtain good results. For example, processing the five more extensive languages of the OSCAR corpus,  English, Russian, Chinese, German, and French, would take around 84 hours in our modest infrastructure. More information can be found in Table \ref{time_multilingual} about the estimated prediction time of different languages using the methodology of the third approach.

\section{Conclusion}
This project proposes an efficient method for labeling harmful content from large volumes of multilingual data extracted from the Internet. We experimented with three approaches, and each approach was built up based on the previous one, aiming to improve the performance step by step. The project's first and second approaches used various classification methods. Different datasets were used, and classification models were created to classify the OSCAR corpus data into harmful and non-harmful content. Both classical machine learning models, transformer-based models and the FastText classifier are convincing in terms of performance, measured by their F1 macro score. However, not all models could be used for classifying OSCAR data due to the computational resources. On top of that, the resulting distribution of the predicted harmful label in the OSCAR corpus made by these models did not present a realistic structure. The experiment used in procedure three showed promising results; the method combined the generation of a language model from Harmful Data from OSCAR and the perplexity score to determine if a text was harmful. In addition to an outstanding classification performance, a realistic distribution of predictions in the OSCAR corpus was achieved. In future works, more data should be generated to train the model. Second, more sophisticated methods to select the perplexity threshold could be applied, such as trying a broader range of threshold values. Third, the methodology of the third approach lays the groundwork for extending its use to other languages. As the solution we proposed here is language independent, the ability to extend this project to a multilingual solution is the most exciting next step.

\bibliography{custom}
\appendix

\section{Appendix}
\label{sec:appendix}

\begin{table}[!ht]
  \centering\small
  \begin{tabular}{lrr}
    \toprule
    Language   & Size     & \begin{tabular}[c]{@{}c@{}}Estimated Time\\  (h)\end{tabular} \\ \midrule
    English    & 3.2 TB   & 44                                                            \\
    Russian    & 1.1 TB   & 15                                                            \\
    Chinese    & 900.9 GB & 7                                                             \\
    German     & 496.7 GB & 13                                                            \\
    French     & 382.2 GB & 5                                                             \\
    Spanish    & 381.9 GB & 5                                                             \\
    Japanese   & 258.7 GB & 4                                                             \\
    Italian    & 229.3 GB & 3                                                             \\
    Portuguese & 170.3 GB & 2                                                             \\
    Dutch      & 114.0 GB & 2                                                             \\ \bottomrule
  \end{tabular}
  \caption{Estimated labeling time per language}
  \label{time_multilingual}
\end{table}

\begin{table*}[!ht]
  \centering\small
  \begin{tabular}{lrclrclr}
    \toprule
    Language   & Size & ~ & Language    & Size & ~ & Language        & Size \\ \midrule
    English    & 2.6G & ~ & Korean      & 4.8M & ~ & Punjabi         & 119K \\
    Swedish    & 647M & ~ & Indonesian  & 4.5M & ~ & Sinhala         & 118K \\
    Russian    & 314M & ~ & Nepali      & 3.4M & ~ & Slovenian       & 107K \\
    French     & 156M & ~ & Kannada     & 2.5M & ~ & Norwegian       & 88K  \\
    Japanese   & 124M & ~ & Danish      & 2.4M & ~ & Telugu          & 78K  \\
    Vietnamese & 112M & ~ & Ukrainian   & 1.9M & ~ & Burmese         & 55K  \\
    Portuguese & 104M & ~ & Finnish     & 1.9M & ~ & Tamil           & 52K  \\
    Italian    & 81M  & ~ & Serbian     & 1.8M & ~ & Tatar           & 47K  \\
    German     & 77M  & ~ & Romanian    & 1.7M & ~ & Mongolian       & 33K  \\
    Chinese    & 74M  & ~ & Albanian    & 1.6M & ~ & Belarusian      & 27K  \\
    Spanish    & 61M  & ~ & Filipino    & 988K & ~ & Egyptian Arabic & 22K  \\
    Dutch      & 61M  & ~ & Latvian     & 705K & ~ & Kazakh          & 20K  \\
    Thai       & 48M  & ~ & Georgian    & 568K & ~ & Tajik           & 20K  \\
    Arabic     & 35M  & ~ & Armenian    & 512K & ~ & Yiddish         & 19K  \\
    Polish     & 26M  & ~ & Hebrew      & 509K & ~ & Macedonian      & 18K  \\
    Turkish    & 17M  & ~ & Gujarati    & 464K & ~ & Sakha           & 17K  \\
    Persian    & 16M  & ~ & Marathi     & 441K & ~ & Amharic         & 11K  \\
    Greek      & 14M  & ~ & Icelandic   & 325K & ~ & Esperanto       & 7.4K \\
    Slovak     & 11M  & ~ & Azerbaijani & 308K & ~ & Latin           & 1.2K \\
    Bangla     & 11M  & ~ & Catalan     & 289K & ~ & Occitan         & 1.1K \\
    Czech      & 9.0M & ~ & Malayalam   & 182K & ~ & Uzbek           & 1.2K \\
    Bulgarian  & 8.8M & ~ & Lao         & 176K & ~ & Sanskrit        & 939  \\
    Hungarian  & 8.5M & ~ & Urdu        & 171K & ~ & ~               & ~    \\
    Hindi      & 7.6M & ~ & Estonian    & 163K & ~ & ~               & ~    \\
    Lithuanian & 5.1M & ~ & Basque      & 140K                              \\ \bottomrule
  \end{tabular}
  \caption{Sizes of adult-tagged data in OSCAR 22.01 by language}
  \label{size_adult}
\end{table*}

\end{document}